# Majorization Minimization Technique for Optimally Solving Deep Dictionary Learning

Vanika Singhal[*], Angshul Majumdar[*]

[*]*Indraprastha Institute of Information Technology, Delhi 110020, India*

Corresponding Author:

Angshul Majumdar,

e-mail: angshul@iiitd.ac.in

tel: +911126907451

The concept of deep dictionary learning has been recently proposed. Unlike shallow dictionary learning which learns single level of dictionary to represent the data, it uses multiple layers of dictionaries. So far, the problem could only be solved in a greedy fashion; this was achieved by learning a single layer of dictionary in each stage where the coefficients from the previous layer acted as inputs to the subsequent layer (only the first layer used the training samples as inputs). This was not optimal; there was feedback from shallower to deeper layers but not the other way. This work proposes an optimal solution to deep dictionary learning whereby all the layers of dictionaries are solved simultaneously. We employ the Majorization Minimization approach. Experiments have been carried out on benchmark datasets; it shows that optimal learning indeed improves over greedy piecemeal learning. Comparison with other unsupervised deep learning tools (stacked denoising autoencoder, deep belief network, contractive autoencoder and K-sparse autoencoder) show that our method supersedes their performance both in accuracy and speed.

*Deep Learning, Dictionary Learning, Optimization*

Glossary of terms when relevant

# Introduction

Today success of deep learning extends beyond academic circles into public knowledge. Perhaps it is the most influential machine learning paradigm of the last decade. Dictionary learning / sparse coding on the other hand enjoyed success, but only within the realms of academia. The recently proposed 'deep dictionary learning' (DDL) [1] combines these two representation learning frameworks.

Dictionary learning is a synthesis representation learning approach; it learns a dictionary so that it can generate / synthesize the data from the learned coefficients. This is a shallow approach – learning only one level of dictionary. Deep dictionary learning extends it to multiple levels. The technique has been proposed in [1]; thorough experimentation [2] showed that it performs better than other unsupervised representation learning tools like stacked denoising autoencoder (SDAE) and deep belief network (DBN). DDL showed promise in an application in hyperspectral imaging [3]; where it was able to show that it significantly surpasses other deep learning techniques when training samples are limited.

However the solution to deep dictionary learning has been far from optimal; it has a greedy solution. In the first level, the dictionary and the coefficients are learnt from the training data as input. In subsequent levels, the coefficients from the previous level acts as input to dictionary learning. Therefore deeper layers are influenced by shallower ones, but not vice versa. Deep learning also follows a greedy learning paradigm, but the issue of feedback from deeper to shallower layers is resolved during the fine-tuning stage.

In this work we propose to rectify this issue; we will learn all the levels of dictionary (and the coefficients) in one optimization problem. However we will not be following the heuristic greedy pre-training followed by fine-tuning paradigm usually employed in deep learning. Our solution will be mathematically elegant. The entire deep dictionary learning problem will be solved in one go using the Majorimization Minimization approach.

The rest of the paper will be organized into several sections. Deep dictionary learning and its relationship with other deep learning tools will be discussed in the following section. Our proposed solution is derived in section 3. Experimental results will be shown in section 4. Finally the conclusions of this work and future direction of research will be discussed in section 5.

## Background

### Representation Learning

Although deep learning has its roots in neural networks, today its reach extends well beyond simple classification. What is more profound is its abstract representation learning ability.

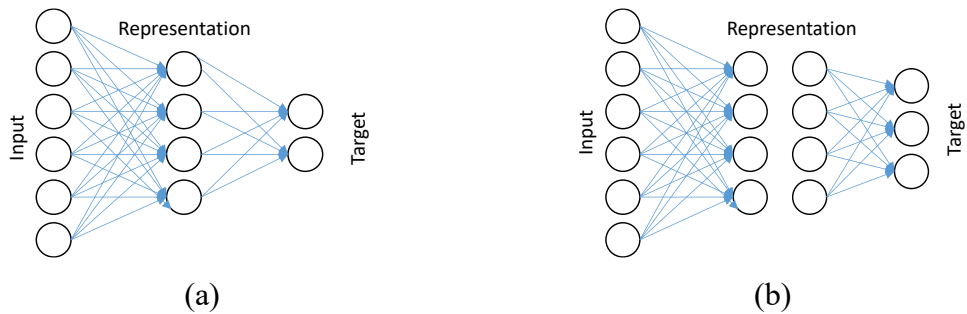

(a)                                        (b)

Fig. 1. (a) Single Representation Layer Neural Network. (b) Segregated Neural Network

Fig. 1(a) shows the diagram of a simple neural network with one representation (hidden) layer. The problem is to learn the network weights between the input and the representation and between the representation and the target. This can be thought of as a segregated problem, see Fig. 1(b). Learning the mapping between the representation and the target is straightforward. This is because once the representation is known, solving for the network weights between the hidden layer and the output target boils down to a simple non-linear least squares problem. The challenge is to learn the network weights (from input) and the representation; this is because we need to solve two variables from one input. Broadly speaking this is the topic of representation learning.

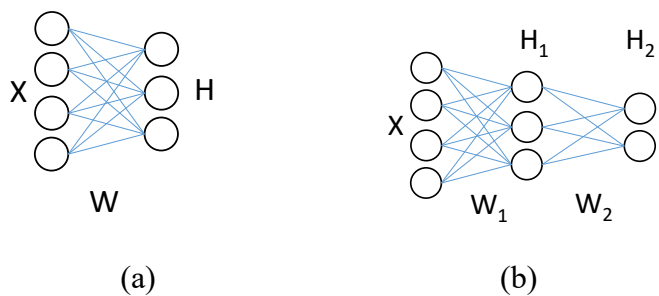

(a)                                        (b)

Fig. 3. (a) Restricted Boltzmann Machine. (b) Deep Boltzmann Machine

Restricted Boltzmann Machine (RBM) [4] is one technique to learn the representation layer. The objective is to learn the network weights ($W$) and the representation ($H$). This is achieved by optimizing the Boltzman cost function given by:

$$p(W, H) = e^{H^T W X} \tag{1}$$

Basically RBM learns the network weights and the representation / feature by maximizing the similarity between the projection of the input (on the network) and the features in a probabilistic sense. Since the usual constraints of probability apply, degenerate solutions are prevented. The traditional RBM is restrictive – it can handle only binary data. The Gaussian-Bernoulli RBM [5] partially overcomes this limitation and can handle real values between 0 and 1. However, it cannot handle arbitrary valued inputs (real or complex).

Deep Boltzmann Machines (DBM) [6, 7] is an extension of RBM by stacking multiple hidden layers on top of each other (Fig. 2(b)). The RBM and DBM are undirected graphical models. These are unsupervised representation learning techniques. For training a deep neural network, targets are attached to the final layer and fine-tuned with back propagation.

The other prevalent technique to train the representation layer of a neural network is by autoencoder [8, 9]. The architecture is shown in Fig. 3(a).

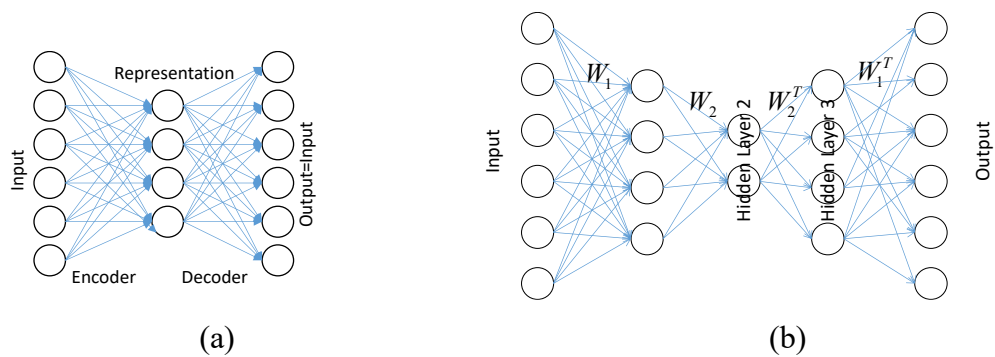

(a)                                                    (b)

Fig. 5. (a) Autoencoder. (b) Stacked Autoencoder

$$\min_{W, W'} \left\| X - W' \phi(WX) \right\|_F^2 \tag{2}$$

The cost function for the autoencoder is expressed above. $W$ is the encoder, and $W'$ is the decoder. The activation function $\varphi$ is usually of tanh or sigmoid such that it squashes the input to normalized values (between 0 and 1 or -1 and +1).

The autoencoder learns the encoder and decoder weights such that the reconstruction error is minimized. Essentially it learns the weights so that the representation $\phi(WX)$ retains almost all the information (in the Euclidean sense) of the data, so that it can be reconstructed back. Once the autoencoder is learnt, the decoder portion of the autoencoder is removed and the target is attached after the representation layer.

To learn multiple layers of representation, the autoencoders are nested into one another. This architecture is called stacked autoencoder, see Fig. 3(b). For such a stacked autoencoder, the optimization problem is complicated. For a two-layer stacked autoencoder, the formulation is,

$$\min_{W_1, W_2, W_1', W_2'} \left\| X - W_1' \varphi \left( W_2' \varphi \left( W_2 \varphi \left( W_1 X \right) \right) \right) \right\|_F^2 \tag{3}$$

The workaround is to learn the layers in a greedy fashion [10]. First the outer layers are learnt (see Fig. 4); and using the features from the outer layer as input for the inner layer, the encoding and decoding weights for the inner layer are learnt.

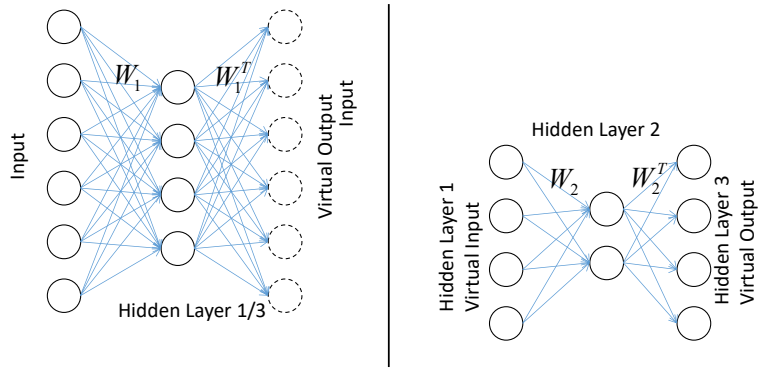

Fig. 6. Greedy Learning

For training deep neural networks, the decoder portion is removed and targets attached to the innermost endoder layer. The complete structure is fine-tuned with backpropagation.

## Deep Dictionary Learning

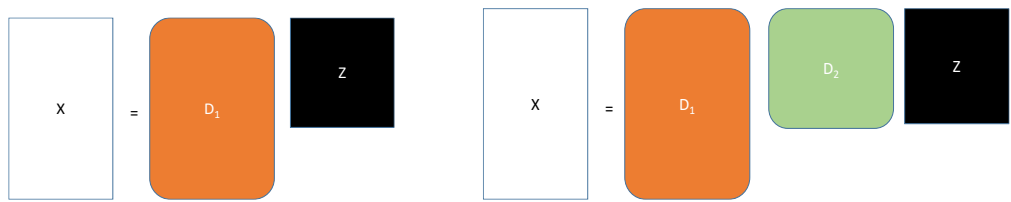

Fig. 7. (a)  Dictionary Learning. (b) – Deep Dictionary Learning

The standard interpretation of dictionary learning is shown in Fig. 1(a). Given the data ($X$), one learns a dictionary $D_1$ so as to synthesize the data from the learnt coefficients $Z$. Mathematically this is expressed as,

$$X = D_1 Z \tag{4}$$

There are several versions of supervised dictionary learning for machine learning application [11, 12]. However in this work we are only interested in the unsupervised version.

In deep dictionary learning, the idea is to learn multiple levels of dictionaries. Deep dictionary learning proposes to extend the shallow dictionary learning into multiple layers – leading to deep dictionary learning, see Fig. 1(b). Mathematically, the representation at the second layer can be written as:

$$X = D_1 \varphi(D_2 Z_2) \tag{5}$$

Extending this idea, a multi-level dictionary learning problem with non-linear activation can be expressed as,

$$X = D_1 \varphi\big( D_2 \varphi(...\varphi(D_N Z))\big) \tag{6}$$

In dictionary learning one usually employs a sparsity penalty on the coefficients. This is required for solving inverse problems [13]; but there is no reason (theoretical or intuitive) for adding the sparsity penalty for learning problems. The seminal paper that started dictionary learning [14], did not impose any sparsity penalty. Without the sparsity penalty, deep dictionary learning leads to,

$$\min_{D_1,...D_N, Z} \big\| X - D_1 \varphi\big( D_2 \varphi(...\varphi(D_N Z))\big)\big\|_F^2 \tag{7}$$

This problem is highly non-convex and requires solving huge number of parameters. With limited amount of data, it will lead to over-fitting. To address these issues, a greedy approach is followed [1-3]. With the substitution $Z_1 = \varphi\big(D_2\varphi(...\varphi(D_N Z))\big)$, Equation (2) can be written as as $X = D_1 Z_1$ such that it can be solved as single layer dictionary learning.

$$\min_{D_1,Z_1} \big\| X - D_1 Z_1\big\|_F^2 \tag{8}$$

This is solved using the method of optimal directions (MOD) [15].

For the second layer, one substitutes $Z_2 = \varphi(D_3...\varphi(D_N Z))$, which leads to $Z_1 = \varphi(D_2 Z_2)$, or alternately, $\varphi^{-1}(Z_1) = D_2 Z_2$; this too is a single layer dictionary learning that can be solved using MOD

$$\min_{D_2,Z_2}\left\|\varphi^{-1}(Z_1)-D_2Z_2\right\|_F^2 \tag{9}$$

Continuing in a similar fashion till the final layer one has

$Z_{N-1}=\varphi(D_NZ)$ or $\varphi^{-1}(Z_{N-1})=D_NZ$. As before, the final level of dictionary and coefficients can be solved using MOD.

This concludes the training stage. During testing, one uses the learnt multi-level dictionaries to generate the coefficients from the test sample. Mathematically one needs to solve,

$$\min_{z_{test}}\left\|x_{test}-D_1\varphi\left(D_2\varphi(...\varphi(D_Nz_{test}))\right)\right\|_2^2 \tag{10}$$

Using the substitution $z_1=\varphi\left(D_2\varphi(...\varphi(D_Nz_{test}))\right)$, learning the feature from the first layer turns out to be,

$$\min_{z_1}\left\|x_{test}-D_1z_1\right\|_2^2 \tag{12}$$

This has a simple analytic solution in the form of pseudoinverse.

With the substitution $Z_2=\varphi(D_3...\varphi(D_NZ))$, one can generate the features at the second level by solving,

$$\min_{z_2}\left\|z_1-\varphi(D_2z_2)\right\|_2^2\equiv\min_{z_2}\left\|\varphi^{-1}(z_1)-D_2z_2\right\|_2^2 \tag{13}$$

The equivalent form has a closed form solution as well. Continuing in this fashion till the final layer, one has

$$\min_{z_{test}}\left\|z_{N-1}-\varphi(D_Nz_{test})\right\|_2^2\equiv\min_{z_{test}}\left\|\varphi^{-1}(z_{N-1})-D_Nz_{test}\right\|_2^2 \tag{14}$$

One can note that the test phase is not very time consuming. One can precompute all the pseudoinverse dictionaries for each level; and can multiply the inputs (after applying inverse of the activation wherever necessary) by these pseudoinverses. Thus during testing, one just needs to compute some matrix vector products; this is the same as any other deep learning tool in test phase.

It must be noted that greedy deep dictionary learning is not the same as deep matrix factorization [16]. Deep matrix factorization is a special case of DDL; where the activations functions are linear. For deep matrix factorization, owing to the linearity of the activation functions one may combine all the levels into a single one; this would collapse the entire deep structure into an equivalent shallow one.

## Proposed Optimal Algorithm

Our goal is to solve (7). Prior studies on deep dictionary learning were only able to solve it greedily in a sub-optimal fashion. For the sake of convenience the problem is repeated.

$$\min_{D_1,...D_N,Z} \left\| X - D_1\varphi\left(D_2\varphi(...\varphi(D_N Z))\right) \right\|_F^2$$

This will be solved using a Majorization Minimization approach. The general outline is discussed in the next sub-section. This is a popular technique in signal processing [17, 18] and machine learning [19, 20], but to the best of our knowledge they have not been used for solving deep learning problems.

### Majorization Minimization

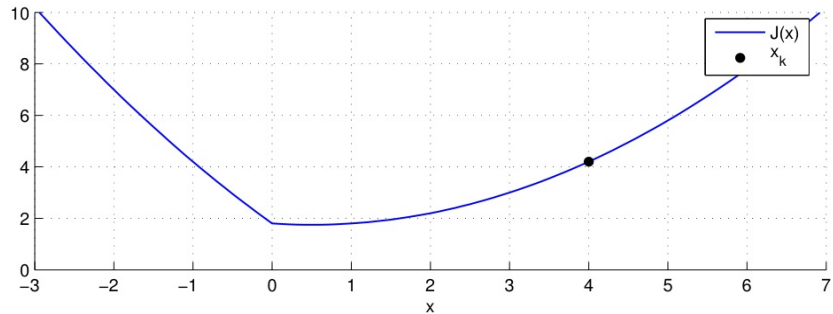

(a)

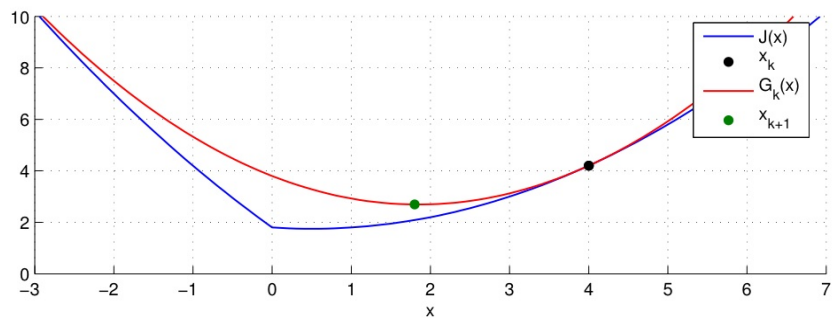

(b)

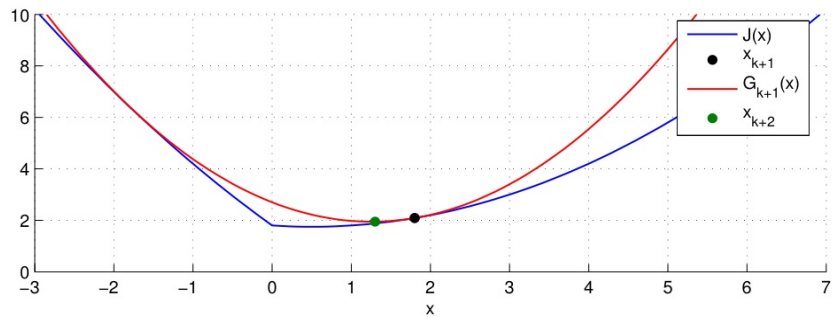

(c)

Fig. 8. Majorization Minmization

Fig. 1 shows the geometrical interpretation behind the Majorization-Minimization (MM) approach. The figure depicts the solution path for a simple scalar problem but essentially captures the MM idea.

Let, J(x) is the function to be minimized. Start with an initial point (at k=0) $x_k$ (Fig. 1a). A smooth function $G_k(x)$ is constructed through $x_k$ which has a higher value than J(x) for all values of x apart from $x_k$, at which the values are the same. This is the Majorization step. The function $G_k(x)$ is constructed such that it is smooth and easy to minimize. At each step, minimize $G_k(x)$ to obtain the next iterate $x_{k+1}$ (Fig 1b). A new $G_{k+1}(x)$ is constructed through $x_{k+1}$ which is now minimized to obtain the next iterate $x_{k+2}$(Fig. 1c). As can be seen, the solution at every iteration gets closer to the actual solution.

## Algorithm Derivation

We will follow an alternating minimization technique for solving the multiple levels of dictionaries and for the final level of coefficients. In every iteration we need to solve for $N$ dictionaries and final level of coefficients $Z$.

For the first level of dictionary, we need to solve,

$$\min_{D_1} \left\| X - D_1 \varphi \left( D_2 \varphi(...\varphi(D_N Z)) \right) \right\|_F^2 \tag{15}$$

Here it is assumed that the dictionaries $D_2$ to $D_N$ and $Z$ are constant while updating $D_1$. For our convenience, we can express $Z_1 = \varphi \left( D_2 \varphi(...\varphi(D_N Z)) \right)$. Thus (15) can be written as,

$$\min_{D_1} \left\| X - D_1 Z_1 \right\|_F^2 \tag{16}$$

One does not need Majorization Minimization to solve this. This (16) is a simple least squares problem with a closed form solution.

Once we have solved $D_1$, we need to solve $D_2$, i.e.

$$\min_{D_2} \left\| X - D_1 \varphi \left( D_2 \varphi(...\varphi(D_N Z)) \right) \right\|_F^2 \tag{17}$$

Expressing $Z_2 = \varphi(D_3...\varphi(D_N Z))$, we get

$$\min_{D_2} \left\| X - D_1 \varphi \left( D_2 Z_2 \right) \right\|_F^2 \tag{18}$$

We need applying Majorization Minimization from now on. Here

$J(D_2) = \left\| X - D_1\varphi(D_2 Z_2) \right\|_F^2$. The majorizer for this (in $k^{th}$ iteration) will be,

$G_k(D_2) = \left\| X - D_1\varphi(D_2 Z_2) \right\|_F^2 + (D_2 - \varphi(D_2 Z_2)_k)^T (aI - D_1^T D_1)(D_2 - \varphi(D_2 Z_2)_k)$

$= X^T X - 2X^T D_1\varphi(D_2 Z_2) + \varphi(D_2 Z_2)^T D_1^T D_1 \varphi(D_2 Z_2)$

$+ (\varphi(D_2 Z_2) - \varphi(D_2 Z_2)_k)^T (aI - D_1^T D_1)(\varphi(D_2 Z_2) - \varphi(D_2 Z_2)_k)$

$= X^T X + \varphi(D_2 Z_2)^T (aI - D_1^T D_1)\varphi(D_2 Z_2)_k - 2(X^T D_1 + x_k^T(aI - D_1^T D_1))\varphi(D_2 Z_2)$

$+ a\varphi(D_2 Z_2)^T \varphi(D_2 Z_2)$

$= a(-2B_1^T D_1 - D_1^T D_1) + c$

where $B_1 = \varphi(D_2 Z_2)_k + \dfrac{1}{a}D_1^T(X - D_1\varphi(D_2 Z_2)_k)$;

$c = X^T X + \varphi(D_2 Z_2)_k^T (aI - D_1^T D_1)\varphi(D_2 Z_2)_k$ and $a$ is the maximum Eigenvalue of $D_1^T D_1$.

Using the identity $\left\| X - Y \right\|_2^2 = X^T X - 2X^T Y + Y^T Y$, one can write,

$$G_k(D_2) = a\left\| B_1 - \varphi(D_2 Z_2) \right\|_F^2 - aB_1^T B_1 + c \qquad (19)$$

Therefore, minimizing (19) is the same as minimizing the first term leaving aside the constants independent of the variable ($D_2$). Therefore, one can instead minimize

$$G_k^{'}(D_2) = \left\| B_1 - \varphi(D_2 Z_2) \right\|_F^2 \qquad (20)$$

where $B_1 = \varphi(D_2 Z_2)_k + \dfrac{1}{a}D_1^T(X - D_1\varphi(D_2 Z_2)_k)$.

Now (20) can be equivalently expressed as,

$$\min_{D_2} \left\| \varphi^{-1}(B_1) - D_2 Z_2 \right\|_F^2 \qquad (21)$$

Computing $\varphi^{-1}$ is easy since it is an elementwise operation. This (21) is a simple least squares solution since $Z_2$ is a constant; as mentioned several times before it has an analytic solution. This concludes the update for $D_2$.

The same technique is continued till deeper layers. For example, solving $D_3$ would require expressing $Z_3 = \varphi(D_4...\varphi(D_N Z))$.

Expanding $Z_2$ in $G_k^{'}(D_2)$ leads to,

$$\left\| \varphi^{-1}(B_1) - D_2\varphi(D_3...\varphi(D_N Z)) \right\|_F^2 \qquad (22)$$

Now, substituting $Z_3 = \varphi(D_4...\varphi(D_N Z))$ in (22) leads to,

$$\min_{D_3} \left\| \varphi^{-1}(B_1) - D_2 \varphi(D_3 Z_3) \right\|_F^2 \tag{24}$$

Note that the problem (24) is exactly the same as (18). Majorization Minimization of (24) leads to

$$\min_{D_3} \left\| B_2 - \varphi(D_3 Z_3) \right\|_F^2 \tag{20}$$

where $B_2 = \varphi(D_3 Z_3)_k + \dfrac{1}{a'} D_2^T (\varphi^{-1}(B_1) - D_2 \varphi(D_3 Z_3)_k)$ ; $a'$ being the maximum eigenvalue of $D_2^T D_2$ .

As before, solving $D_3$ from the equivalent expression $\min_{D_3} \left\| \varphi^{-1}(B_2) - D_3 Z_3 \right\|_F^2$ is straightforward.

We continue this till the pre-final layer; after solving $D_{N-1}$ we are left with the solution of the final level of dictionary $D_N$ coefficients $Z$. Majorization Minimization would lead to an expression similar to (20); we will have

$$\min_{D_N, Z} \left\| B_{N-1} - \varphi(D_N Z) \right\|_F^2 \tag{21}$$

Unlike the other layers, we can solve for both the dictionary and the coefficients of the final layer by simple alternating least squares (ALS) / MOD of the following equivalent form.

$$\min_{D_N, Z} \left\| \varphi^{-1}(B_{N-1}) - D_N Z \right\|_F^2 \tag{22}$$

The ALS / MOD algorithm is succinctly shown below.

---

Initialize: $D_N$

Update $Z$: $\min_{Z} \left\| \varphi^{-1}(B_{N-1}) - (D_N)_{k-1} Z \right\|_F^2$

Update $D_N$: $\min_{D_N} \left\| \varphi^{-1}(B_{N-1}) - D_N (Z)_k \right\|_F^2$

---

Note that our method is completely non-parametric; therefore there is nothing to tune, once the number of dictionaries and the number of atoms in each are fixed by the user.

Our proposed derivation results in a nested algorithm, i.e. for one update of $D_1$, the update for $D_2$ is in a loop; similarly for one update of $D_2$, the update for $D_3$ is in a loop and so on. Succinctly the algorithm can be expressed as follows:

---

Initialize: $D_2, D_3, \ldots, D_N$ and $Z$.

Loop 1

---

$D_1 \leftarrow \min\limits_{D_1} \|X - D_1 Z_1\|_F^2$ where $Z_1 = \varphi\left(D_2 \varphi(...\varphi(D_N Z))\right)$

  Loop 2

  $D_2 \leftarrow \min\limits_{D_2} \|\varphi^{-1}(B_1) - D_2 Z_2\|_F^2$ where $Z_2 = \varphi(D_3...\varphi(D_N Z))$

and   $B_1 = \varphi\left(D_2 Z_2\right)_k + \dfrac{1}{a} D_1^T (X - D_1 \varphi\left(D_2 Z_2\right)_k)$

    Loop 3

    $\min\limits_{D_3} \|\varphi^{-1}(B_2) - D_3 Z_3\|_F^2$ where $Z_3 = \varphi(D_4...\varphi(D_N Z))$

    and   $B_2 = \varphi\left(D_3 Z_3\right)_k + \dfrac{1}{a'} D_2^T (\varphi^{-1}(B_1) - D_2 \varphi\left(D_3 Z_3\right)_k)$

      Loop 4

      .....

        Loop N

        $Z \leftarrow \min\limits_{Z} \|\varphi^{-1}(B_{N-1}) - (D_N)_{k-1} Z\|_F^2$

        $D_N \leftarrow \min\limits_{D_N} \|\varphi^{-1}(B_{N-1}) - D_N(Z)_k\|_F^2$

        End Loop N

      …

      End Loop 4

    End Loop 3

  End Loop 2

End Loop 1

To prevent degenerate solutions where some of the $D$'s are very high and others low, the columns of all the dictionaries are normalized after every update.

The initialization is done deterministically. First the SVD of $X$ is computed ($X = USV^T$) and $D_1$ is initialized by the top left eigenvectors of $X$. For $D_2$, the SVD of SVD is computed and the corresponding top eigenvectors are used to initialized $D_2$. The rest of the dictionaries are initialized in a similar fashion. In the last level, the coefficient ($Z$) is initialized by the product of the eigenvalues and the right eigenvectors of the last SVD. There can be other randomized techniques for initialization which may yield better results, but our deterministic initialization is repeatable and has shown to yield good results consistently.

For our proposed algorithm ideally one needs to run the loops for several iterations. This would be very time consuming; we found that in practice it is not required. Only the deepest loop for updating $D_N$ and $Z$ is solved for 5 to 10 iterations. The rest of the loops from 2 to N-1 are only run once. Only the outermost loop is run for a large number of iterations (~100).

There will be no variation in the testing phase. As discussed before, once the dictionaries are learnt, the feature generation during testing is fast – one only needs a few (equaling the number of levels) matrix vector multiplication.

## Experimental Evaluation

### Datasets

We carried our experiments on several benchmarks datasets. The first one is the MNIST dataset which consists of 28x28 images of handwritten digits ranging from 0 to 9. The dataset has 60,000 images for training and 10,000 images for testing. No preprocessing has been done on this dataset.

We also tested on variations of MNIST, which are more challenging primarily because they have fewer training samples (10,000 + 2,000 validation) and larger number of test samples (50,000). This one was created specifically to benchmark deep learning algorithms [21].

1. basic (smaller subset of MNIST)
2. basic-rot (smaller subset with random rotations)
3. bg-rand (smaller subset with uniformly distributed noise in background)
4. bg-img (smaller subset with random image background)
5. bg-img-rot (smaller subset with random image background plus rotation)

We have also evaluated on the problem of classifying documents into their corresponding newsgroup topic. We have used a version of the 20-newsgroup dataset [22] for which the training and test sets contain documents collected at different times, a setting that is more reflective of a practical application. The training set consists of 11,269 samples and the test set contains 7,505 examples. We have used 5000 most frequent words for the binary input features. We follow the same protocol as outlined in [23].

Our third dataset is the GTZAN music genre dataset [24, 25]. The dataset contains 10000 three-second audio clips, equally distributed among 10 musical genres:

blues, classical, country, disco, hip-hop, pop, jazz, metal, reggae and rock.  Each example in the set is represented by 592 Mel-Phon Coefficient (MPC) features. These are a simplified formulation of the Mel-frequency Cepstral Coefficients (MFCCs) that are shown to yield better classification performance. Since there is no predefined standard split and fewer examples, we have used 10-fold cross validation (procedure mentioned in [26]), where each fold consisted of 9000 training examples and 1000 test examples.

## Results

In this work our goal is to test the representation capability of the different learning tools. Therefore the training is fully unsupervised (no class label is used). We compare against several state-of-the-art unsupervised deep learning tools – stacked denoising autoencoder (SDAE) [26], K-sparse autoencoder (KSAE) [27], Contractive Autoencoder (CAE) [28] and Deep Belief Network [29]. Since our goal is to show that our proposed optimal learning algorithm yields improvement over the greedy deep dictionary learning technique proposed before [1-3], we carry out comparison with this as well. Learned models for the popular datasets used in this work are publicly available. For the deep dictionary learning (previous [1] and proposed), a three layer architecture is used where the number of atoms are halved in each subsequent layer.

The generated features from the deepest level are used to train two non-parametric – nearest neighbor (NN) (Table 1) and sparse representation based classification (SRC) [30] (Table 2); and one parametric – support vector machine (SVM) classifier with rbf kernel (Table 3). The results show that our proposed method yields the best results on an average.

Table 1. Comparison on KNN

| Dataset | SDAE | KSAE | CAE | DBN | Greedy DDL | Proposed |
|---|---|---|---|---|---|---|
| MNIST | 97.33 | 96.90 | 92.83 | 97.05 | 97.75 | **97.91** |
| basic | 95.25 | 91.64 | 90.92 | 95.37 | 95.80 | **96.07** |
| basic-rot | 84.83 | 80.24 | 78.56 | 84.71 | 87.00 | **87.23** |
| bg-rand | 86.42 | 85.89 | 85.61 | 86.36 | 89.35 | **89.77** |
| bg-img | 77.16 | 76.84 | 78.51 | 77.16 | 81.00 | **81.09** |
| bg-img-rot | 52.21 | 50.27 | 47.10 | 50.47 | 57.77 | **58.40** |
| 20-newsgroup | 70.48 | 71.22 | 71.08 | 70.09 | 70.48 | **71.64** |

| GTZAN | 83.31 | 82.91 | 82.67 | 80.99 | 83.31 | **83.89** |

Table 2. Comparison on SRC

| Dataset | SDAE | KSAE | CAE | DBN | Greedy DDL | Proposed |
|---|---|---|---|---|---|---|
| MNIST | **98.33** | 97.91 | 87.19 | 88.43 | 97.99 | **98.33** |
| basic | 96.91 | 95.07 | 95.03 | 87.49 | 96.38 | **96.97** |
| basic-rot | 90.04 | 88.85 | 88.63 | 79.47 | 89.74 | **90.23** |
| bg-rand | 91.03 | 83.59 | 82.25 | 79.67 | 91.38 | **91.61** |
| bg-img | 84.14 | 84.12 | 85.68 | 75.09 | 84.11 | 84.67 |
| bg-img-rot | 62.46 | 58.06 | 54.01 | 49.68 | 62.86 | **63.27** |
| 20-newsgroup | 70.49 | 71.90 | 71.08 | 71.02 | 71.41 | **72.43** |
| GTZAN | 83.37 | 84.09 | 82.70 | 81.21 | 84.72 | **85.71** |

Table 3. Comparison on SVM

| Dataset | SDAE | KSAE | CAE | DBN | Greedy DDL | Proposed |
|---|---|---|---|---|---|---|
| MNIST | 98.50 | 98.46 | 97.74 | 98.53 | 98.64 | **98.71** |
| basic | 96.96 | 97.02 | 96.61 | 97.07 | 97.28 | **97.53** |
| basic-rot | 89.43 | 88.75 | 72.54 | 89.05 | 90.34 | **90.75** |
| bg-rand | 91.28 | 90.07 | 85.20 | 89.59 | 92.38 | **92.62** |
| bg-img | 84.86 | 80.17 | 78.76 | 85.46 | 86.17 | **86.67** |
| bg-img-rot | 60.53 | 60.01 | 60.97 | 58.25 | 63.85 | **64.76** |
| 20-newsgroup | 71.29 | 72.05 | 71.68 | 71.18 | 71.97 | **72.89** |
| GTZAN | 83.42 | 81.61 | 82.99 | 81.83 | 84.92 | **85.18** |

The results are as expected. In the prior studies [1, 2] it was already shown that greedy DDL outperforms SDAE and DBN. We now see that, it also improves upon K-sparse autoencoder and contractive autoencoder.

Since this is a new (optimal) algorithm for solving the unsupervised deep dictionary learning problem, we need to test its speed. The training and testing times for the large MNIST dataset and the relatively smaller MNIST basic dataset are shown in Tables 4 and 5. All the algorithms are run until convergence on a machine with Intel (R) Core(TM) $i5$ running at 3 GHz; 8 GB RAM, Windows 10 (64 bit) running Matlab 2014a.

Table 4. Training Time in Seconds

| Dataset | SDAE | KSAE | CAE | DBN | Greedy DDL | Proposed |
|---|---|---|---|---|---|---|

| MNIST | 120408 | 59251 | 40980 | 30071 | 107 | 524 |
| basic | 24020 | 10031 | 8290 | 5974 | 26 | 129 |

Table 5. Testing Time in Seconds

| Dataset | SDAE | KSAE | CAE | DBN | Greedy DDL | Proposed |
|---------|------|------|-----|-----|-----------|----------|
| MNIST | 61 | 52 | 56 | 50 | 79 | 51 |
| basic | 257 | 206 | 214 | 155 | 189 | 189 |

The training time of our proposed algorithm is significantly larger than the greedy approach; this is expected. But still we are significantly faster, by several orders of magnitude, compared to other deep learning tools. In terms of testing time, we are faster than greedy deep dictionary learning. This is because the greedy technique uses standard dictionary learning tools in each level; these are always regularized by sparsity promoting penalties on the coefficients. Thus during testing, one needs to solve an iterative optimization problem. Our formulation on the other hand does not include sparsity promoting $l_1/l_0$-norm; hence each level can be solved via an analytic solution (pseudoinverse). Therefore we just need a matrix vector multiplication. Hence we take almost the same time as other deep learning tools while testing.

## Conclusion

A new deep learning tool called deep dictionary learning has been recently proposed. The idea there is to represent the training data as a non-linear combination of several layers of dictionaries. All prior studies were only able to solve the ensuing problem in a greedy fashion. This was a sub-optimal solution as there was no flow of information from the deeper to the shallower layers. This is the first work that proposes an optimal solution to the deep dictionary learning problem; where all the levels of dictionaries are solved simultaneously as a single optimization problem. We invoke the Majorization Minimization framework to solve the said problem. This results in an algorithm that is completely non-parametric.

Experiments have been carried out on several benchmark datasets. In all of them, our method performs the best. The only downside of our algorithm (compared to the existing greedy technique) is that ours is comparatively slower than greedy deep dictionary learning. Nevertheless, we are still several orders of magnitude faster than other unsupervised deep learning tools.

## Acknowledgement

The authors thank the Infosys Center for Artificial Intelligence for partial support.